\pdfoutput=1
\documentclass[letterpaper]{article}
\usepackage{iccc}

\usepackage{xurl}
\usepackage{times}
\usepackage{helvet}
\usepackage{courier}
\usepackage{soul,color}
\usepackage{subcaption}
\usepackage{graphicx}
\usepackage{natbib}
\usepackage{flushend} % Makes both columns equal length on last page
\title{Active Divergence with Generative Deep Learning - A Survey and Taxonomy}
\author{
  Terence Broad\,\textsuperscript{1,\,2}{\rm,}
  Sebastian Berns\,\textsuperscript{3}{\rm,}
  Simon Colton\,\textsuperscript{3,\,4} \and
  Mick Grierson\,\textsuperscript{2} \\
  \textsuperscript{1}\,Department of Computing, Goldsmiths, University of London, UK \\
  \textsuperscript{2}\,Creative Computing Institute, University of The Arts London, UK \\
  \textsuperscript{3}\,School of Electronic Engineering and Computer Science, Queen Mary University of London, UK \\
  \textsuperscript{4}\,SensiLab, Faculty of IT, Monash University, Melbourne, Australia \\

}
\begin{document} 
\maketitle
\begin{abstract}
\begin{quote}
Generative deep learning systems offer powerful tools for artefact generation, given their ability to model distributions of data and generate high-fidelity results. In the context of computational creativity, however, a major shortcoming is that they are unable to explicitly diverge from the training data in creative ways and are limited to fitting the target data distribution. To address these limitations, there have been a growing number of approaches for optimising, hacking and rewriting these models in order to actively diverge from the training data. We present a taxonomy and comprehensive survey of the state of the art of \emph{active divergence} techniques, highlighting the potential for computational creativity researchers to advance these methods and use deep generative models in truly creative systems.

\end{quote}
\end{abstract}

\section{Introduction}

Generative deep learning methods, and in particular deep generative models, have become very powerful at producing high quality artefacts and have garnered a huge amount of interest in machine learning, computer graphics and audio signal processing communities. In addition, because they are capable of producing artefacts of high cultural value, they are also of interest to artists and for the development of creativity support tools. 

One of the main goals of researchers in computational creativity and by artists and others using generative deep learning systems, is to find ways to get generative models to produce novel outcomes that diverge from the training data. In some respects, attempting to create a generative model that does not model the training data is an oxymoron, as by definition a generative \textit{model} must model some existing data distribution. However, generative neural networks are powerful tools with the unique capability of learning to render entire distributions of complex high dimensional data with ever-increasing fidelity. It is no wonder then, that there have been a large number of approaches developed in order tweak, manipulate and optimise these models in order to actively diverge from the training data, or any existing data distribution.

The term \textit{active divergence} \citep{berns2020bridging} describes methods for utilising generative deep learning in ways that do not simply reproduce the training data. Methods for this have been developed within the field of computational creativity, but also a goal commonly shared by neighbouring communities, such as those building creativity support tools and artists, researchers and other pracitioners publishing and sharing results under the `CreativeAI' banner \citep{cook2018neighbouring}. This paper offers a comprehensive survey and taxonomy of the state of the art with respect to methods developed across these fields.

Additionally, this paper outlines some of the possible applications, and outlines key opportunities for computational creativity research to advance active divergence methods beyond tricks and hacks, towards more automated and autonomous creative systems. Many of the research directions presented are still very nascent and a lot of work is still to be done in regards to evaluating and benchmarking these methods. Better ways of measuring and evaluating these techniques will go a long way to advancing understanding and allowing more creative responsibility to be handed over to the systems. The comparative account of the methods, use-cases and future research directions for active divergence is offered as a resource to inform future research in generative deep learning tools and systems that take creative leaps beyond reproducing the training data.

\section{Technical Overview}

While not all generative models rely on generative deep learning, we refer here to those that build on artificial neural networks\footnote{For further reading, a comprehensive overview of generative models is given in \citet{harshvardhan2020comprehensive}.}. Given a data distribution $P$, a generative model will model an approximate distribution $P'$. The parameters for the approximate distribution can be learned by an artificial neural network. This learning task is tackled differently by different architectures and training schemes. E.g. autoencoders \citep{rumelhart1985learning} and variational autoencoders (VAE) \citep{kingma2013auto,rezende2014stochastic} learn to approximate the data through reconstruction via an encoding and a decoding network, while generative adversarial networks (GAN) \citep{goodfellow2014generative} consists of a generator that is guided by a discriminating network. In most cases, the network learns a mapping from a lower-dimensional latent distribution $X$ to the complex high-dimensional feature space of a domain. The model, thus, generates a sample $p'$ given an input vector $x$ which should resemble samples drawn from the target distribution $P$. In the simplest case of a one layer network the generated sample $p'$ is generated using the function: $p' = \sigma(Wx+b)$ where $x$ is the input vector from the latent distribution $x \in X$, $\sigma$ is a non-linear activation function, $W$ and $b$ are the learned association matrix and bias vector for generating samples in the approximate distribution $p' \in P'$. The model parameters $W$ and $b$, are typically learned through gradient-based optimisation process. In this process, a loss function will require the model to maximise the likelihood of the data either: (i) explicitly, as in the case of autoencoders, autoregressive \citep{frey1996does} and flow-based generative models \citep{dinh2014nice}; (ii) approximately, as is the case in VAEs; (iii) or implicitly, as in the case of GANs. Generative models can also be conditioned on labelled data. In the conditional case, the generative model takes two inputs $x$ and $y$, where $y$ represents the class label vector. Another form of conditional generative models are translation models, such as pix2pix \citep{isola2017image}, that takes a (high dimensional) data distribution as input $Q$ and learns a mapping to $P'$ which is an approximation of the true target function $f: Q \rightarrow P$.

All deep generative models, and in particular ones that generate high dimensional data domains like images, audio and natural language, will have some level of divergence $D(P||P') \geq 0$ between the target distribution $P$ and the approximate distribution $P'$, because of the complexity and stochasticity inherent in high dimensional data. The goal of all generative models is to minimise that level of divergence, by maximising the likelihood of generating the given data domain. Active divergence methods however, intentionally seek to create a new distribution $U$ that does not directly approximate a given distribution $P$, or resemble any other known data distribution. This is either done by seeking to find model parameters $W^*$ and $b^*$ (in the single layer case) that generate novel samples $u = \sigma(W^*x+b^*)$, or by making other kinds of interventions to the chain of computations.

\section{Survey of Active Divergence Methods}

We present a comprehensive overview and taxonomy of the state of the art in methods for achieving active divergence. In this survey, we will use the term divergence in the statistical sense, as being the distance (or difference) between two distributions. There are other definitions of divergence relevant to research in creativity, such as Guildford's dimensions of divergent thought \citep{hocevar1980intelligence}. While there are some parallels that can be drawn between some of the active divergence methods, and theories of divergent thinking; for the clarity of technical exposition, we will be sticking strictly to the statistical definition of divergence in this overview of active divergence methods. 

% These methods have been advanced by researchers in computational creativity and deep learning, as well as artists, researchers and other practitioners who advance techniques under the `CreativeAI' banner. 

\begin{figure*}[tbp]
   \centering
   \begin{subfigure}[b]{0.245\textwidth}
     \centering
     \includegraphics[width=\textwidth]{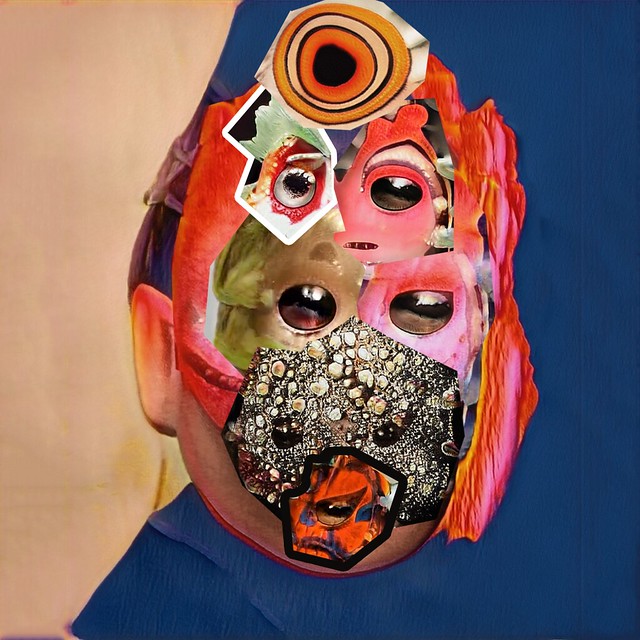}
     \caption{divergent fine-tuning }
     \label{fig:fruits}
   \end{subfigure}
   \hfill
   \begin{subfigure}[b]{0.245\textwidth}
     \centering
     \includegraphics[width=\textwidth]{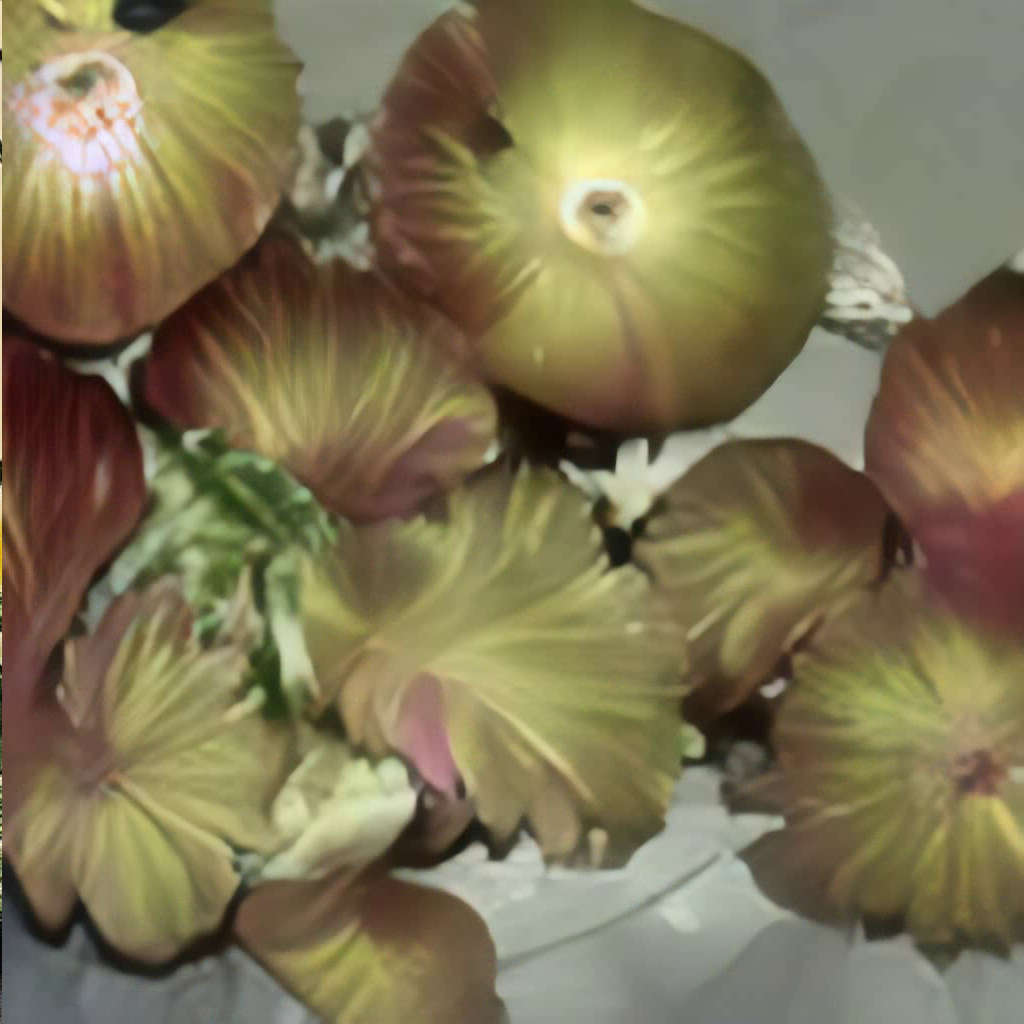}
     \caption{chaining models}
     \label{fig:chaining}
   \end{subfigure}
   \begin{subfigure}[b]{0.245\textwidth}
     \centering
     \includegraphics[width=\textwidth]{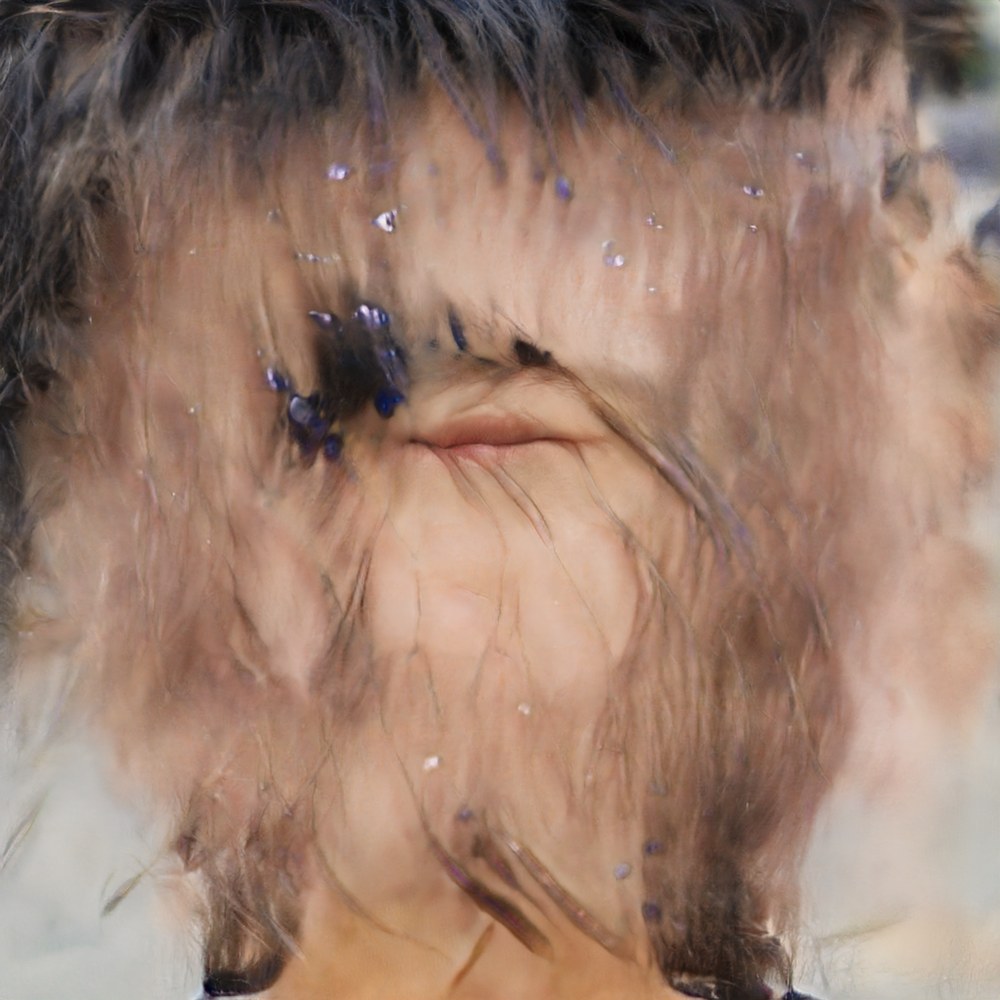}
     \caption{network bending}
     \label{fig:teratome}
   \end{subfigure}
   \hfill
   \begin{subfigure}[b]{0.245\textwidth}
     \centering
     \includegraphics[width=\textwidth]{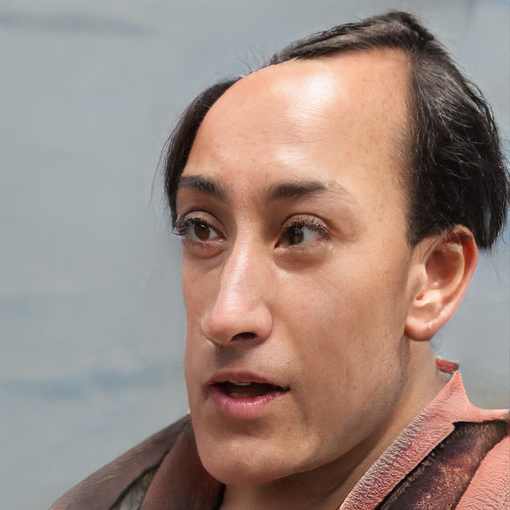}
     \caption{network blending}
     \label{fig:layerswap}
   \end{subfigure}
    \caption{Some visual examples of results produced using various active divergence methods. (a) An image from \textit{Strange Fruit} by Mal Som~\protect\citep{som2020strange}, that was created by fine-tuning a pre-trained model towards a continously shifting domain. (b)~A frame from the video artwork \textit{You Are Here} by Derrick Schultz~\protect\citep{schutlz2020you}, created by chaining multiple models and technniques including: a custom GAN, network bending, image translation, and super-resolution. (c)~An image from the series of artworks \textit{Teratome}~\protect\citep{broad2020teratome}, that was created using network bending techniques~\protect\citep{broad2021network}. (d)~An example of network blending~\protect\citep{pinkney2020interpolation},  where the image provided has been generated from a model which combines the photorealistic textures from the FFHQ StyleGAN2 model, but the spatial structure from a model trained on an Ukiyo-e dataset~\protect\citep{pinkney2020aligned}. All images are reproduced with permission from their respective creators.}
    \label{fig:three graphs}
\end{figure*}

\subsection{Novelty search over learned representations}
\label{survey:noveltysearch}

Methods in this category take existing generative models trained using standard maximum likelihood regimes and then specifically search for the subset of learned representations that do not resemble the training data by systematically sampling from the model\footnote{An overview of methods for sampling generative models is given in \citet{white2016sampling}.}. Taking account of the fact that any approximate distribution $P'$ will be somewhat divergent from the true distribution $P$, these methods seek to find the subset $U$ of the approximate distribution which is not contained in the true distribution $U \subset P' \wedge U \not\subset P$. \citet{kazakcci2016digits} present an algorithm for searching for novelty in the latent space of a sparse autoencoder trained on the MNIST dataset \citep{lecun1998gradient}. They start by creating a sample of random noise and by using a Markov chain monte carlo (MCMC) method of iteratively re-encoding the sample through the encoder, then refining the sample until it produces a stable representation. They use this approach to map out all the representations the model can generate, then perform k-means clustering on the latent space encoding of these representations. By disregarding clusters that correspond to real digits, they are left with clusters of representations of digits that do not exist in the original data distribution. It has been argued that these `spurious samples' are the inevitable outcome of generative models that learn to generalise from given data distributions \citep{kegl2018spurious} and that there is a trade off between the ability to generalise to every mode in the dataset and the ratio of spurious samples in the resulting distribution.

\subsection{Novelty generation from an inspiring set}
\label{survey:noveltygeneration}

The methods in this section train a model from scratch using a training dataset, but do not attempt to model the data directly, rather using it as reference material to draw inspiration from. We therefore refer to this training set (the given distribution $P$) as the inspiring set \citep{ritchie2007some}.

An approach for novel glyph generation utilises a class-conditional generative model trained on the MNIST dataset \citep{lecun1998gradient}, but in this case they train the model with `hold-out classes' \citep{cherti2017out}, additional classes that do not exist in the training data distribution. These hold-out classes can then sampled during inference, which encapsulate the subset $U$ of the approximate distribution $P'$ that is not included in the target distribution $U \subset P' \wedge U \not\subset P$. These divergent samples can then be generated directly by conditioning the generator with the hold-out class label, without the need for searching the latent space. 

An approach that directly generates a new distribution $U$ from an inspiring set $P$ is the creative adversarial networks (CAN) algorithm \citep{elgammal2017can}. The algorithm uses the WikiArt dataset \citep{saleh2016large}, a labelled dataset of paintings classified by `style' (historical art movement). This algorithm draws inspiration from the GAN training procedure \citep{goodfellow2014generative}, but adapts it such that the discriminator has to classify real and generated samples by style, and the generator is then optimised to maximise the likelihood of the generated results being classified as `artworks' (samples that fit the training distribution of existing artworks) but maximise their deviation from existing styles in order to produce the novel distribution $U$.

\subsection{Training without data}
\label{survey:nodata}

Training a model from a random initial starting point without any training data, almost certainly guarantees novelty in the resulting generated distribution. Existing approaches to doing this all rely on the dynamics between multiple models to produce emergent behaviours through which novel data distributions can be generated. 

\subsubsection{Multi-generator dynamics} \citet{broad2019searching} present an approach to training generative deep learning models without any training data, by using two generator networks, and relying on the dynamics between them for an open-ended optimisation process. This approach took inspiration from the GAN framework, but instead of a generator mimicking real data, two generators attempt to mimic each other while the discriminator attempts to tell them apart. In order to have some level of diversity in the final results, the two generators are simultaneously trying to produce more colours in the generated output than the other generator network, leading to the generation of two novel, yet closely related distributions $U$ and $V$.

\subsubsection{Generation via communication} An alternative approach to generating without data uses a single generator network, and uses the generated distribution $U$ as a channel for communication between two networks, which together learn to generate and classify images that represent numerical and textual information from a range of existing datasets \citep{simon2019dimensions}. In subsequent work, by constraining the generator with a strong inductive bias for generating line drawings, this approach can be utilised for novel glyph generation \citep{park2020generating}.

\subsection{Divergent fine-tuning}
\label{survey:divergent}

Divergent fine-tuning methods take pre-trained models that generate an approximate distribution $P'$ and fine-tune the model away from the original training data. This can either be done by optimising on new training data, or by using auxiliary models and custom loss functions. The goal being to find a new set of model parameters that generate a novel distribution $U$, that is significantly divergent from the approximate distribution $P'$ and the original distribution $P$.

\subsubsection{Cross domain training} In cross domain training, transfer learning is performed to a pre-trained model that generates the approximate distribution $P'$ and is then trained to approximate the new data distribution $Q$. This transfer learning procedure will eventually lead to the model learning a set of parameters that generate the approximate distribution $Q'$. However, by picking an iteration of the model mid-way through this process, a set of parameters can be found that produced a blend between the two approximate distributions $P'$ and $Q'$, resulting in the producing the novel distribution $U$ \citep{schultz2020mixed}. This method, was discovered by many artists and practitioners independently, who were performing transfer learning with GAN models for training efficiency, but noted that the iterations of the model part-way through produced the most interesting, surprising and sometimes horrifying results \citep{adler2020transfer,black2020noface,mariansky2020transfer,shane2020cat}.

\subsubsection{Continual domain shift}
\label{section:domainshift}
Going beyond simply mixing two domains, one approach that gives more opportunity to steer the resulting distribution in the fine-tuning procedure, is to optimise on a domain that is continually shifting. In creating the artworks \textit{Strange Fruit} \citep{som2020strange}, the artist Mal Som ``iterate[s] on the dataset with augmenting, duplicating and looping in generated images from previous ticks'' to steer the training of the generator model \citep{som2021personal}. In this process, the target distribution $Q_t$ at step $t$ may contain samples $q'_{t-n}$ generated from earlier iterations of the model at any previous time step $t-n$ where $0<n<t$. Additionally, the target distribution $Q_t$, may no longer include samples, or may have duplicates of samples $q_{t-n}$ from previous iterations of the target distribution. Using this process, the target distribution can be continually shaped and guided. 

This process of modelling a continually shifting domain often leads to the ---generally unwanted--- phenomenon of mode collapse \citep{thanh2020catastrophic}. However, in Som's practice, this is induced deliberately. After a model has collapsed, Som explores its previous iterations to find the last usable instance right before collapse. Som likens this practice to the artistic technique of defamiliarisation, where common things are presented in unfamiliar ways so audiences can gain new perspectives and see the world differently \citep{som2021personal}.

\subsubsection{Loss hacking} An alternative strategy, is to fine-tune a model without any training data. Instead a loss function is used that directly transforms the approximate distribution $P'$ into a novel distribution $U$ without requiring any other target distribution. \citet{broad2020amplifying} use the frozen weights of the discriminator to directly optimise away from the likelihood of the data, by using the inverse of the adversarial loss function. This process reverses the normal objective of the generator to generate `real' data and instead to generate samples that the discriminator deems to be `fake'. By applying this process to a GAN that can produce photo-realistic images of faces, this fine-tuning procedure crosses the uncanny valley in reverse, taking images indistinguishable from real images, and amplifying the uncanniness of the images before eventually leading to mode collapse. In a similar fashion to Som's practice (see previous sub-section), one instance of the model before mode collapse was hand-selected and a selection of its outputs turned into the series of artworks \textit{Being Foiled} \citep{broad2020being}.

\subsubsection{Infusing external knowledge} 

By harnessing the learned knowledge of externally trained models, it is possible to fine-tune models to infuse that knowledge to transform the original domain data with characteristics defined using the auxiliary model. \citet{broad2019transforming} utilise a classifier model $C_{classifier}$ trained to differentiate between datasets, in conjunction with the frozen weights of the discriminator $D_{frozen}$ to fine-tune a pre-trained GAN generator model $G$ away from the original distribution and towards a new local minimum defined by the loss function $L$. $L$ is defined as the weighted sum of the two auxiliary models $L = \alpha C_{classifier}(G(x)) + \beta D_{frozen}(G(x))$ given the random latent vector $x$, and $\alpha$ and $\beta$ being the hyper-parameters defining the weightings for the two components of the loss function. 

The StyleGAN-NADA framework \citep{gal2021stylegan} takes advantage of the external knowledge of a contrastive language–image pre-training model (CLIP) \citep{radford2021learning}. CLIP has been trained on billions of text and image pairs from the internet and provides a joint-embedding space of both images and text, allowing for similarity estimation of images and text prompts. In StyleGAN-NADA, pretrained StyleGAN2 models \citep{karras2019analyzing} can be fine-tuned using user-specified text prompts, the CLIP model $C_{clip}$ is then used to encode the text prompts and the generated samples in order to provide a loss function where the cosine similarity $S$ between the clip encodings of the text string $t$ and the generated image embedding $G(x)$ given random latent $x$, can be minimised using the loss $L = S(C_{clip}(t), C_{clip}(G(x))$. This training procedure, guides the generator towards infusing characteristics from an unseen domain defined by the user as text prompts.

\subsection{Chaining models}
\label{survey:chaining}

An approach that is widely used by artists who incorporate generative models into their practice, but not well documented in academic literature, is the practice of chaining multiple custom models trained on datasets curated by the artists. The ensembles used will often utilise standard unconditional generative models, such as GANs, in combination with other conditional generative models such as image-to-image translation networks, such as pix2pix \citep{isola2017image} and CycleGAN \citep{zhu2017unpaired}, along with other approaches for altering the aesthetic outcomes of results such as style transfer \citep{gatys2016neural}. Artists will often train many models on small custom datasets and test out many combinations of different models, with the aim of finding a configuration that produces unique and expressive results. The artist Helena Sarin will often chain multiple CycleGAN models into one ensemble, and will reuse training data during inference, as the goal of this practice ``is not generalization, my goal is to create appealing art'' \citep{sarin2018playing}. The artist Derrick Schultz draws parallels between the practice of chaining models and Robin Sloan's concept of `flip-flopping' \citep{schultz2021personal}, where creative outcomes can be achieved by ``pushing a work of art or craft from the physical world to the digital world and back, often more than once'' \citep{sloan2012flipflop}.

\subsection{Network bending}
\label{survey:bending}

Network bending \citep{broad2021network} is a framework that allows for active divergence using individual pre-trained models without making any changes to the weights or topology of the model. Instead, additional layers that implement standard image filters are inserted into the computational graph of a model and applied during inference to the activation maps of the convolutional features\footnote{Inserting filters into GANs was also developed independently in the Matlab StyleGAN playground \citep{pinkney2020matlab}.}. As the computational graph of the model has been altered, the model which previously generated samples from the approximate distribution $P'$, now produces novel samples from the new distribution $U$, without any changes being made to the parameters of the model. In the simplest case of a two layer model an association weight matrix $W_l$ and bias $b_l$ vector for each layer $l$. Which generates sample $p'=\sigma(W_2(\sigma(W_1x+b_1))+b_2)$ from input vector $x$ and using a non-linear activation function $\sigma$. In the network bending framework, a deterministic function $f$ (controlled by the parameter $y$) is inserted into the computational graph of the model and applied to the internal activations of the model $u=\sigma(W_2(f(\sigma(W_1x+b_1),y))+b_2)$, allowing the model to produce new samples $u$ from the new distribution $u \in U$. Beyond the simplest case of a transformation being applied to all features in a layer, the transformation layer can also be applied to a random sub-section of features, or to a pre-selected set of features. \citet{broad2021network} present a clustering algorithm, that in an unsupervised fashion, groups together sets of features within a layer based on the spatial similarity of their activation maps. This clustering algorithm is capable of finding sets of features responsible for the generation of various semantically meaningful components of the generated output across the network (and semantic) hierarchy, which can then be manipulated in tandem allowing for semantic manipulation of the internal representations of the generative model. 

In addition to applying filters to the activation maps, it is also possible to enlarge samples by increasing the size of the activation maps and interpolating and tiling them \citep{pouliot2020gan}. The network bending framework has been extended into the domain of audio synthesis \citep{mccallum2020network} where it has been applied to neural vocoder models using the differential digital signal processing (DDSP) approach \citep{engel2020ddsp}. In order to adapt the framework for the audio domain, \citet{mccallum2020network} implement a number of filters that operate in the time domain, such as oscillators. Network bending has also been applied in the domain of audio-reactive visual synthesis using generative models \citep{brouwer2020audio}, with the deterministic transformations being controlled automatically using features extracted from audio analysis.

\subsection{Network blending}
\label{survey:blending}

Blending multiple models trained on different dataset allows for more control over the combination of learned features from different domains. This can either be done by blending the predictions of the models, or by blending the parameters of the models themselves.

\subsubsection{Blending model predictions} \citet{akten2016real} present an interactive tool for text generation allowing for the realtime blending of the predicted outputs of an ensemble of long-short term memory network (LSTM) models \citep{hochreiter1997long} trained to perform next character prediction from different text sources. A graphical user interface allows the user to dynamically shift the mixture weights for the weighted sum for the predictions of all of the models in the ensemble, prior to the one hot vector encoding which is used to determine the final predicted character value.

\subsubsection{Blending model parameters} A number of approaches, all demonstrated with StyleGAN2 \citep{karras2019analyzing}, take advantage of the large number of pre-trained models that have been shared on the internet \citep{pinkney2020awesome}. Of these almost all have been transfer-learned from the official model weights trained on the Flickr-Faces High Quality (FFHQ) dataset. It has been shown that the parameters of models transfer-learned $p_{transfer}$ from the same original source $p_{base}$ share commonalities in the way their weights are structured. This makes it possible to meaningfully interpolate between the parameters of the models directly \citep{aydao2020interp}. By using an interpolation weighting $\alpha$, it is possible to control the interpolation for the creation of a set of parameters $p_{interp} = (1 - \alpha)p_{base} + \alpha p_{transfer}$. 

Layers can also be swapped from one model to another \citep{pinkney2020interpolation}, allowing the combination of higher level features of one model with lower level features of another. This layer swapping technique was used to make the popular `toonification' method, which can be used to find the corresponding sample to a real photograph of a person in a Disney-Pixar-esque `toonified' model, simply by sampling from the same latent vector that has been found as the closest match to the person in FFHQ latent space \citep{abdal2019image2stylegan}. A generalised approach that combines both weight interpolation and layer-swapping methods for multiple models, uses a cascade of different weightings of interpolation for the various layers of the model \citep{arfafax2020barycentricnotebook}.

\citet{colton2021evolving} presents an evolutionary approach for exploring and finding effective and customisable neural style transfer blends. Upwards of 1000 neural style transfer models trained on 1-10 style images each, can be blended through model interpolation, using an interface that is controlled by the user. MAP-Elites \citep{mouret2015illuminating} in combination with a fitness function calculated using the output from a ResNet model \citep{he2016deep} were used in evolutionary searches for optimal neural style transfer blends. 

\subsection{Model rewriting}
\label{survey:rewriting}

Model rewriting encompasses approaches where either the weights or network topology are altered in a targeted way, through manual intervention or by using some form of heuristic based optimisation algorithm. 

\subsubsection{Stochastic rewriting} To create the series of artworks \textit{Neural Glitch} the artist Mario Klingemann randomly altered, deleteed or exchanged the trained weights of pre-trained GANs \citep{klingemann2018neural}. In a similar fashion, the convolutional layer reconnection technique \citep{ruzika2020gan} randomly swaps convolutional features within layers of pre-trained GANs. This technique is applied in the \textit{Remixing AIs} audiovisual synthesis framework \citep{collins2020remixing}.

\subsubsection{Targeted rewriting} \citet{bau2020rewriting} present a targeted approach to model rewriting. Here, a sample is taken from the model and manipulated using standard image editing techniques (referred to as a `copy-paste' interface). Once the sample has been altered corresponding to the desired goal (such as removing watermarks from the image, or getting horses to wear hats), a process of constrained optimisation is performed. All of the layers but one are frozen, and the weights of that layer are updated using gradient descent optimisation until the generated sample matches the new target. After this optimisation process is complete, the weights of the model are modified such that the targeted change becomes present in all the samples that the model generates.

The CombiNets framework \citep{guzdial2018combinets}, informed by prior reseach in combinational creativity \citep{boden2004creative}, can be utilised to create a new model by combining parameters from a number of pre-trained models in a targeted fashion. The parameters of existing models are recombined to take into account a new mode of generation that was not present in the training data (an example given would be a unicorn for a model trained on photographs of non-mythical beings). In this framework, a small number of new samples is provided (not enough to train a model directly) and then heuristic search is used to recombine parameters from existing models to account for this new mode of generation.

\section{Further Demarcations}

In this section, we highlight demarcations that can be used to classify methods for active divergence. The following categories serve as criteria for further discussion and method comparison.

\subsection{Training from scratch vs. using pre-trained models}

Finding stable, effective ways of training generative models, in particular GANs, is difficult and, depending on the training scheme, there are only a handful of methods that have been found to work successfully. Few methods for active divergence train a model completely from scratch. Instead, most take pre-trained models as their starting point for interventions. This way, training from scratch can be avoided, but fine-tuning may still be required.

\subsection{Utilising data vs. dataless approaches}
Most of the approaches described utilise data in some way, whether as an inspiring set for novelty generation, or for combining features from different datasets (divergent fine-tuning, network blending and chaining models). Even methods for model rewriting use very small amounts of example data to guide optimisation algorithms that alter the model weights. However, methods like network bending, show how models can be analysed in ways that don't rely on any data, and are used for intelligent manipulation of the models ---an approach which could be applied to other methods like model rewriting. Methods that train and fine-tune models without data also show how auxiliary networks and the dynamics between models can be utilised for achieving active divergence.

\subsection{Human direction vs. creative autonomy}
Very few of the approaches described have been developed with the expressed intention of handing over creative agency to the systems themselves. Most of the methods have been developed by artists or researchers in order to allow people to manipulate, experiment with and explore the unintended uses of these models for creative expression. However, the methods described that are currently designed for, or rely on a high degree of human curation and intervention, could easily be adapted and used in co-creative or autonomous creative systems in the future \citep{berns2021automating}.

\section{Applications of Active Divergence}

In this section we outline some of the applications for active divergence methods. 

\subsection{Novelty generation}

Generative deep learning techniques are capable of generalisation, such that they can produce new artefacts of high typicality and value, but are rarely capable of producing novel outputs that do not resemble the training data. Active divergence techniques play an important role in getting generative deep learning systems to generate truly novel artefacts, especially when there may be limited or even no data to draw from. 

\subsection{Creativity support and co-creation}

Some of the frameworks presented are already explicitly designed as creativity support tools, such as the network bending framework, designed to allow for expressive manipulation of deep generative models. The \textit{Style Done Quick} \citep{colton2021evolving} application where many style transfer models have been evolved, was built as a casual creator application \citep{compton2015casual}. Though many of the other methods described are still preliminary artistic and research experiments, there is a lot of potential for these methods to become better understood and eventually adapted and applied in more easily accessible creativity support tools and co-creation frameworks. 

\subsection{Knowledge recombination}

Reusing and recombining knowledge in efficient ways is an important use-case of active divergence methods. While impressive generalisation can be ascertained from extremely large models trained on corpuses extracted from large portions of the internet \citep{ramesh2021zero}, this is out of the capabilities for all but a handful of large tech companies. Instead of relying on ever expanding computational resources, active divergence methods allow for the recombination of styles, aesthetic characteristics and higher level concepts in a much more efficient fashion. Methods like chaining models, network blending and model rewriting offer alternatives routes to achieving flexible knowledge recombination and generalisation to unseen domains without the need for extremely large models or data sources.

\subsection{Unseen domain adaptation}

Active divergence methods allow for the possibility of adapting to and exploring unseen domains, for which there is little to no data available. The network blending approach presented by \citet{pinkney2020interpolation} can be used for the translation of faces while maintaining recognisable identity into a completely synthesised data domain, something which would not be possible with standard techniques for image translation \citep{zhu2017unpaired}.

The model rewriting and network bending approaches offer the possibility of reusing and manipulating existing knowledge in a controlled fashion to create new data from a small number of given examples, or theoretically without any prior examples if external knowledge sources are integrated, as discussed further below. This approach could also be utilised by agents looking to explore hypothetical situations, by reorganising learned knowledge from world models \citep{ha2018worldmodels} to explore hypothetical situations or relations.

\subsection{A benchmark for creativity}

Generative models represent large knowledge bases that can produce high quality artefacts. There is a lot of unexplored potential for how the information and relationships they contain can be reused and rewritten with frameworks for manipulating them such as network bending and model re-writing. Active divergence frameworks could make good candidates for exploring and evaluating modes of creativity, such as combinational creativity \citep{boden2004creative} and conceptual blending \citep{fauconnier2008way}. These could be used to inform how the features in the model could be re-organised, and then evaluated by examining the artefacts generated from the altered models. 

\section{Future Research Directions}

In this section we discuss possible future research directions and applications for developing, evaluating and utilising methods for active divergence.

\subsection{Metrics for quantitative evaluation}

For the advancment of research on active divergence, methods for quantitative evaluation will be critical in order to keep track of progress, to compare techniques and for benchmarking. Metrics for active divergence will have to go beyond measuring the similarity or dissimilarity between distributions, as is usually done in the evaluation of generative models \citep{gretton2019interpretable}. Active divergence metrics should contribute to a better understanding of \textit{how} the distributions diverge. Therefore, various changes to the modelled distribution should be taken into consideration when looking to measure divergence between distributions in creative contexts. These include increases or decreases in diversity, the consistency and concurrency of change across the whole distribution and whether changes primarily effect low or high level features.

\subsection{Automating qualitative evaluation}

In addition to quantitative evaluation, other metrics are needed for evaluating active divergence metrics. In order to rely less on qualitative evaluation for guiding decisions in creating new models, and do this in computational fashion so that these aspects of the process can be handed over to the computational systems. For instance, a recently developed metric for measuring visual indeterminacy \citep{wang2020towards}, which is argued as being one of the key drivers for what people find interesting in GAN generated art \citep{hertzmann2020visual}, could be used for replacing the qualitative evaluation and curation step done by humans. Other metrics that could be used are: novelty metrics \citep{grace2019expectation}, bayesian surprise \citep{itti2009bayesian}, aesthetic evaluation \citep{galanter2012computational}, or measurements for optimal blends between data domains and evaluating the novelty of changes made to semantic relationships.

\subsection{Inventing new objective functions}

None of the methods presented to date that are based on generative deep learning have been capable of inventing their own objective functions. Instead, methods such as creative adversarial networks \citep{elgammal2017can} rely on hand crafted variations of well established objective functions. This will be one of most challenging future research directions to overcome, as generative deep learning systems rely on a small handful of objectives that result in stable convergence. However, in conjunction with the development of new evaluation metrics, it may be possible to explore whole new categories of objective functions that diverge from existing data representations and produce artefacts of high-value. 

\subsection{Utilising external knowledge}
\label{future:external}
Harnessing expert knowledge external to the dataset, which may come from separate domains or symbolic knowledge representations will allow much more flexibility in how generative models are manipulated in combinational creativity \citep{boden2004creative} and conceptual blending frameworks \citep{fauconnier2008way}. Combining research into analysing the semantic purpose and relationship between features, and creating mappings of those to external data sources or knowledge graphs, would allow for more flexibility in controlling techniques which currently rely on human intervention (network bending, model rewriting). This could be adapted to be controlled and manipulated computationally, allowing for some creative decision making to be handed over to the computer.

\subsection{Formulating and realising intentions}

For many of the methods described, a system that could formulate and realise its intentions would have to be capable of sourcing and creating its own dataset. For instance, a system that wants to create a model that generates hybrids between cats and dogs, would have to be capable of collecting data of cats and dogs separately, and then decide to use some method for network blending to get the desired results. Alternatively, utilising external knowledge sources in combination with semantic analysis of features, would allow computational systems more flexibility in generating new models by altering the semantic relationships between features in model rewriting or network bending approaches. 

\subsection{Multi-agent systems}

It has been argued the the GAN framework is the simplest example of a multi-agent system \citep{arcas2019social}, and frameworks such as neural cellular automata \citep{mordvintsev2020growing} offer new possibilities for multi-agent approaches in generative deep learning. The active divergence methods for training without data described in this paper all rely on the dynamics of multiple agents to produce interesting results, but this could be taken much further. It has been argued that art is fundamentally social \citep{hertzmann2021social} and exploring more complex social dynamics between agents \citep{saunders2019multi} could be a fruitful avenue for exploration in the development of these approaches. There is a large body of work in emergent languages from co-operative multi-agent systems \citep{lazaridou2016multi} that could be drawn from in furthering the work in generative multi-agent systems. 

\subsection{Open-ended reinforcement learning}

Open-ended reinforcement learning, where there is no set goal \citep{wang2020enhanced}, offers possibilities for new more autonomous approaches to achieving active divergence. Reinforcement learning has not been discussed in this survey, but has been used in generative settings \citep{luo2020reinforcement} in nascent research. Reinforcement learning approaches offer many opportunities for frameworks of creativity to be explored that are not available to standard generative deep learning methods, as they take actions in response to their environment, rather than just fitting functions. Paradigms like intrinsic motivation \citep{shaker2016intrinsically}, cooperating or competing with other agents, formulating and acting on intentions are all concepts that conventional generative deep learning systems alone cannot explore, but these paradigms could be explored in open-ended systems utilising reinforcement learning.

\section{Conclusion}

We have presented a taxonomy and survey of the state of the art in methods for achieving active divergence from a range of sources, including artistic experiments, creativity support tools and in computational creativity research. Many of these methods represent nascent areas of research and there is a lot of scope for future work utilising them in co-creative and automated creative systems as they overcome a key shortcoming of mainstream generative deep learning approaches, which are unable to diverge from reproducing the training data in creative ways. In addition, we outline a number of the key future research directions needed in order to advance the state of the art for creativity support tools and computationally creative generative deep learning systems. 

\subsection{Acknowledgements}
We thank our reviewers for their helpful comments.
This work has been supported by UK’s EPSRC Centre for Doctoral Training in Intelligent Games and Game Intelligence (IGGI; grants EP/L015846/1 and EP/S022325/1).

%\appendix{\LaTeX{} and Word Style Files}\label{stylefiles}

%The \LaTeX{} and Word style files are available on the ICCC-13
%website, {\tt http://computationalcreativity.net/iccc2013/}.
%These style files implement the formatting instructions in this
%document.

%The \LaTeX{} files are {\tt iccc.sty} and {\tt iccc.tex}, and
%the Bib\TeX{} files are {\tt iccc.bst} and {\tt iccc.bib}. The
%\LaTeX{} style file is for version 2e of \LaTeX{}, and the Bib\TeX{}
%style file is for version 0.99c of Bib\TeX{} ({\em not} version
%0.98i).

%The Microsoft Word style file consists of a single template file, {\tt
%iccc.dot}. 

%These Microsoft Word and \LaTeX{} files contain the source of the
%present document and may serve as a formatting sample. 

\bibliographystyle{iccc}
\bibliography{iccc}

\begin{thebibliography}{}

\bibitem[\protect\citeauthoryear{Abdal, Qin, and
  Wonka}{2019}]{abdal2019image2stylegan}
Abdal, R.; Qin, Y.; and Wonka, P.
\newblock 2019.
\newblock {Image2StyleGAN}: How to embed images into the {StyleGAN} latent
  space?
\newblock In {\em IEEE International Conference on Computer Vision},
  4432--4441.

\bibitem[\protect\citeauthoryear{Adler}{2020}]{adler2020transfer}
Adler, D.
\newblock 2020.
\newblock Deliberate stylegan2 ffhq corruption. fine tuned upon a tiny set
  [...].
\newblock \url{https://twitter.com/Norod78/status/1218282356391530496}.
\newblock Accessed: 2021-02-05.

\bibitem[\protect\citeauthoryear{Agüera~y Arcas}{2019}]{arcas2019social}
Agüera~y Arcas, B.
\newblock 2019.
\newblock Social intelligence.
\newblock In {\em Advances in Neural Information Processing Systems [Keynote
  address]}.

\bibitem[\protect\citeauthoryear{Akten and Grierson}{2016}]{akten2016real}
Akten, M., and Grierson, M.
\newblock 2016.
\newblock Real-time interactive sequence generation and control with recurrent
  neural network ensembles.
\newblock {\em Recurrent Neural Networks Symposium, NIPS 2016}.

\bibitem[\protect\citeauthoryear{Arfafax}{2020}]{arfafax2020barycentricnotebook}
Arfafax.
\newblock 2020.
\newblock Barycentric cross-network interpolation with different layer
  interpolation rates.
\newblock
  \url{https://colab.research.google.com/drive/1FwOYqtU0kVYDwHrddFKBhDKcs0jJ_zuK}.
\newblock Accessed: 2020-02-05.

\bibitem[\protect\citeauthoryear{Aydao}{2020}]{aydao2020interp}
Aydao.
\newblock 2020.
\newblock Yeah stochastic weight averaging of neural networks is wild [...].
\newblock \url{https://twitter.com/AydaoAI/status/1234614081413406720}.
\newblock Accessed: 2021-02-05.

\bibitem[\protect\citeauthoryear{Bau \bgroup et al.\egroup
  }{2020}]{bau2020rewriting}
Bau, D.; Liu, S.; Wang, T.; Zhu, J.-Y.; and Torralba, A.
\newblock 2020.
\newblock Rewriting a deep generative model.
\newblock In {\em Proc. European Conference on Computer Vision (ECCV)}.

\bibitem[\protect\citeauthoryear{Berns and Colton}{2020}]{berns2020bridging}
Berns, S., and Colton, S.
\newblock 2020.
\newblock Bridging generative deep learning and computational creativity.
\newblock In {\em Proc. 11th International Conference on Computational
  Creativity}.

\bibitem[\protect\citeauthoryear{Berns \bgroup et al.\egroup
  }{2021}]{berns2021automating}
Berns, S.; Broad, T.; Guckelsberger, C.; and Colton, S.
\newblock 2021.
\newblock {Automating Generative Deep Learning for Artistic Purposes:
  Challenges and Opportunities}.
\newblock In {\em Proc. 12th International Conference on Computational
  Creativity}.

\bibitem[\protect\citeauthoryear{Black}{2020}]{black2020noface}
Black, S.
\newblock 2020.
\newblock Thanks! it's trained on faces then trained a little while [...].
\newblock \url{https://twitter.com/realmeatyhuman/status/1257733313885765638}.
\newblock Accessed: 2021-02-05.

\bibitem[\protect\citeauthoryear{Boden}{2004}]{boden2004creative}
Boden, M.~A.
\newblock 2004.
\newblock {\em The creative mind: Myths and mechanisms}.
\newblock Psychology Press.

\bibitem[\protect\citeauthoryear{Broad and
  Grierson}{2019a}]{broad2019searching}
Broad, T., and Grierson, M.
\newblock 2019a.
\newblock Searching for an (un)stable equilibrium: experiments in training
  generative models without data.
\newblock {\em NeurIPS 2019 Workshop on Machine Learning for Creativity and
  Design}.

\bibitem[\protect\citeauthoryear{Broad and
  Grierson}{2019b}]{broad2019transforming}
Broad, T., and Grierson, M.
\newblock 2019b.
\newblock Transforming the output of {GANs} by fine-tuning them with features
  from different datasets.
\newblock {\em arXiv preprint arXiv:1910.02411}.

\bibitem[\protect\citeauthoryear{Broad, Leymarie, and
  Grierson}{2020}]{broad2020amplifying}
Broad, T.; Leymarie, F.~F.; and Grierson, M.
\newblock 2020.
\newblock Amplifying the uncanny.
\newblock {\em Proc. 8th Conference on Computation, Communication, Aesthetics
  and X (xCoAx)}.

\bibitem[\protect\citeauthoryear{Broad, Leymarie, and
  Grierson}{2021}]{broad2021network}
Broad, T.; Leymarie, F.~F.; and Grierson, M.
\newblock 2021.
\newblock Network bending: Expressive manipulation of deep generative models.
\newblock {\em Proc. 10th International Conference on Artificial Intelligence
  in Music, Sound, Art and Design (EvoMUSART).}

\bibitem[\protect\citeauthoryear{Broad}{2020a}]{broad2020being}
Broad, T.
\newblock 2020a.
\newblock {Being Foiled}.
\newblock \url{https://terencebroad.com/works/being-foiled}.
\newblock Accessed: 2021-06-30.

\bibitem[\protect\citeauthoryear{Broad}{2020b}]{broad2020teratome}
Broad, T.
\newblock 2020b.
\newblock Teratome.
\newblock \url{https://terencebroad.com/works/teratome}.
\newblock Accessed: 2021-06-28.

\bibitem[\protect\citeauthoryear{Brouwer}{2020}]{brouwer2020audio}
Brouwer, H.
\newblock 2020.
\newblock Audio-reactive latent interpolations with {StyleGAN}.
\newblock {\em NeurIPS 2020 Workshop on Machine Learning for Creativity and
  Design}.

\bibitem[\protect\citeauthoryear{Cherti, K{\'e}gl, and
  Kazak{\c{c}}{\i}}{2017}]{cherti2017out}
Cherti, M.; K{\'e}gl, B.; and Kazak{\c{c}}{\i}, A.
\newblock 2017.
\newblock Out-of-class novelty generation: an experimental foundation.
\newblock In {\em Proc. IEEE 29th International Conference on Tools with
  Artificial Intelligence (ICTAI)}.

\bibitem[\protect\citeauthoryear{Collins, Růžička, and
  Grierson}{2020}]{collins2020remixing}
Collins, N.; Růžička, V.; and Grierson, M.
\newblock 2020.
\newblock Remixing ais: mind swaps, hybrainity, and splicing musical models.
\newblock In {\em Proc. The Joint Conference on AI Music Creativity}.

\bibitem[\protect\citeauthoryear{Colton}{2021}]{colton2021evolving}
Colton, S.
\newblock 2021.
\newblock Evolving neural style transfer blends.
\newblock {\em Proc. 10th International Conference on Artificial Intelligence
  in Music, Sound, Art and Design (EvoMUSART).}

\bibitem[\protect\citeauthoryear{Compton and Mateas}{2015}]{compton2015casual}
Compton, K., and Mateas, M.
\newblock 2015.
\newblock Casual creators.
\newblock In {\em Proc. 6th International Conference on Computational
  Creativity}.

\bibitem[\protect\citeauthoryear{Cook and Colton}{2018}]{cook2018neighbouring}
Cook, M., and Colton, S.
\newblock 2018.
\newblock Neighbouring communities: Interaction, lessons and opportunities.
\newblock {\em Association for Computational Creativity (ACC)}.

\bibitem[\protect\citeauthoryear{Dinh, Krueger, and
  Bengio}{2014}]{dinh2014nice}
Dinh, L.; Krueger, D.; and Bengio, Y.
\newblock 2014.
\newblock Nice: Non-linear independent components estimation.
\newblock {\em arXiv preprint arXiv:1410.8516}.

\bibitem[\protect\citeauthoryear{Elgammal \bgroup et al.\egroup
  }{2017}]{elgammal2017can}
Elgammal, A.; Liu, B.; Elhoseiny, M.; and Mazzone, M.
\newblock 2017.
\newblock {CAN}: Creative adversarial networks, generating" art" by learning
  about styles and deviating from style norms.
\newblock {\em Proc. 8th International Conference on Computational Creativity}.

\bibitem[\protect\citeauthoryear{Engel \bgroup et al.\egroup
  }{2020}]{engel2020ddsp}
Engel, J.; Hantrakul, L.; Gu, C.; and Roberts, A.
\newblock 2020.
\newblock {DDSP}: Differentiable digital signal processing.
\newblock {\em International Conference on Learning Representations}.

\bibitem[\protect\citeauthoryear{Fauconnier and
  Turner}{2008}]{fauconnier2008way}
Fauconnier, G., and Turner, M.
\newblock 2008.
\newblock {\em The way we think: Conceptual blending and the mind's hidden
  complexities}.
\newblock Basic Books.

\bibitem[\protect\citeauthoryear{Frey \bgroup et al.\egroup
  }{1996}]{frey1996does}
Frey, B.~J.; Hinton, G.~E.; Dayan, P.; et~al.
\newblock 1996.
\newblock Does the wake-sleep algorithm produce good density estimators?
\newblock In {\em Advances in neural information processing systems},
  661--670.
\newblock Citeseer.

\bibitem[\protect\citeauthoryear{Gal}{2021}]{gal2021stylegan}
Gal, R.
\newblock 2021.
\newblock {StyleGAN2-NADA}.
\newblock \url{https://github.com/rinongal/StyleGAN-nada}.
\newblock Accessed: 2021-06-28.

\bibitem[\protect\citeauthoryear{Galanter}{2012}]{galanter2012computational}
Galanter, P.
\newblock 2012.
\newblock Computational aesthetic evaluation: past and future.
\newblock {\em Computers and creativity}  255--293.

\bibitem[\protect\citeauthoryear{Gatys, Ecker, and
  Bethge}{2016}]{gatys2016neural}
Gatys, L.; Ecker, A.; and Bethge, M.
\newblock 2016.
\newblock A neural algorithm of artistic style.
\newblock {\em Journal of Vision} 16(12):326--326.

\bibitem[\protect\citeauthoryear{Goodfellow \bgroup et al.\egroup
  }{2014}]{goodfellow2014generative}
Goodfellow, I.; Pouget-Abadie, J.; Mirza, M.; Xu, B.; Warde-Farley, D.; Ozair,
  S.; Courville, A.; and Bengio, Y.
\newblock 2014.
\newblock Generative adversarial nets.
\newblock In {\em Advances in neural information processing systems}.

\bibitem[\protect\citeauthoryear{Grace and Maher}{2019}]{grace2019expectation}
Grace, K., and Maher, M.~L.
\newblock 2019.
\newblock Expectation-based models of novelty for evaluating computational
  creativity.
\newblock In {\em Computational Creativity}. Springer.
\newblock  195--209.

\bibitem[\protect\citeauthoryear{Gretton, Sutherland, and
  Jitkrittum}{2019}]{gretton2019interpretable}
Gretton, A.; Sutherland, D.; and Jitkrittum, W.
\newblock 2019.
\newblock Interpretable comparison of distributions and models.
\newblock In {\em Advances in Neural Information Processing Systems
  [Tutorial]}.

\bibitem[\protect\citeauthoryear{Guzdial and
  Riedl}{2018}]{guzdial2018combinets}
Guzdial, M., and Riedl, M.~O.
\newblock 2018.
\newblock Combinets: Creativity via recombination of neural networks.
\newblock {\em Proc. 9th International Conference on Computational Creativity}.

\bibitem[\protect\citeauthoryear{Ha and Schmidhuber}{2018}]{ha2018worldmodels}
Ha, D., and Schmidhuber, J.
\newblock 2018.
\newblock Recurrent world models facilitate policy evolution.
\newblock {\em Advances in Neural Information Processing Systems 31}.

\bibitem[\protect\citeauthoryear{Harshvardhan \bgroup et al.\egroup
  }{2020}]{harshvardhan2020comprehensive}
Harshvardhan, G.; Gourisaria, M.~K.; Pandey, M.; and Rautaray, S.~S.
\newblock 2020.
\newblock A comprehensive survey and analysis of generative models in machine
  learning.
\newblock {\em Computer Science Review} 38:100285.

\bibitem[\protect\citeauthoryear{He \bgroup et al.\egroup }{2016}]{he2016deep}
He, K.; Zhang, X.; Ren, S.; and Sun, J.
\newblock 2016.
\newblock Deep residual learning for image recognition.
\newblock In {\em Proc. IEEE conference on computer vision and pattern
  recognition}.

\bibitem[\protect\citeauthoryear{Hertzmann}{2020}]{hertzmann2020visual}
Hertzmann, A.
\newblock 2020.
\newblock Visual indeterminacy in {GAN} art.
\newblock {\em Leonardo} 53(4):424--428.

\bibitem[\protect\citeauthoryear{Hertzmann}{2021}]{hertzmann2021social}
Hertzmann, A.
\newblock 2021.
\newblock Art is fundamentally social.
\newblock \url{https://aaronhertzmann.com/2021/03/22/art-is-social.html}.
\newblock Accessed: 2020-03-29.

\bibitem[\protect\citeauthoryear{Hocevar}{1980}]{hocevar1980intelligence}
Hocevar, D.
\newblock 1980.
\newblock Intelligence, divergent thinking, and creativity.
\newblock {\em Intelligence} 4(1):25--40.

\bibitem[\protect\citeauthoryear{Hochreiter and
  Schmidhuber}{1997}]{hochreiter1997long}
Hochreiter, S., and Schmidhuber, J.
\newblock 1997.
\newblock Long short-term memory.
\newblock {\em Neural computation} 9(8).

\bibitem[\protect\citeauthoryear{Isola \bgroup et al.\egroup
  }{2017}]{isola2017image}
Isola, P.; Zhu, J.-Y.; Zhou, T.; and Efros, A.~A.
\newblock 2017.
\newblock Image-to-image translation with conditional adversarial networks.
\newblock In {\em Proc. IEEE Conference on Computer Vision and Pattern
  Recognition}.

\bibitem[\protect\citeauthoryear{Itti and Baldi}{2009}]{itti2009bayesian}
Itti, L., and Baldi, P.
\newblock 2009.
\newblock Bayesian surprise attracts human attention.
\newblock {\em Vision research} 49(10):1295--1306.

\bibitem[\protect\citeauthoryear{Karras \bgroup et al.\egroup
  }{2020}]{karras2019analyzing}
Karras, T.; Laine, S.; Aittala, M.; Hellsten, J.; Lehtinen, J.; and Aila, T.
\newblock 2020.
\newblock Analyzing and improving the image quality of {StyleGAN}.
\newblock {\em Proc. IEEE Conference on Computer Vision and Pattern
  Recognition}.

\bibitem[\protect\citeauthoryear{Kazak{\c{c}}{\i}, Mehdi, and
  K{\'e}gl}{2016}]{kazakcci2016digits}
Kazak{\c{c}}{\i}, A.; Mehdi, C.; and K{\'e}gl, B.
\newblock 2016.
\newblock Digits that are not: Generating new types through deep neural nets.
\newblock In {\em Proc. 7th International Conference on Computational
  Creativity}.

\bibitem[\protect\citeauthoryear{K{\'e}gl, Cherti, and
  Kazak{\c{c}}{\i}}{2018}]{kegl2018spurious}
K{\'e}gl, B.; Cherti, M.; and Kazak{\c{c}}{\i}, A.
\newblock 2018.
\newblock Spurious samples in deep generative models: bug or feature?
\newblock {\em arXiv preprint arXiv:1810.01876}.

\bibitem[\protect\citeauthoryear{Kingma and Welling}{2013}]{kingma2013auto}
Kingma, D.~P., and Welling, M.
\newblock 2013.
\newblock Auto-encoding variational {B}ayes.
\newblock In {\em International Conference on Learning Representations}.

\bibitem[\protect\citeauthoryear{Klingemann}{2018}]{klingemann2018neural}
Klingemann, M.
\newblock 2018.
\newblock Neural glitch / mistaken identity.
\newblock \url{https://underdestruction.com/2018/10/28/neural-glitch/}.
\newblock Accessed: 2021-02-05.

\bibitem[\protect\citeauthoryear{Lazaridou, Peysakhovich, and
  Baroni}{2017}]{lazaridou2016multi}
Lazaridou, A.; Peysakhovich, A.; and Baroni, M.
\newblock 2017.
\newblock Multi-agent cooperation and the emergence of (natural) language.
\newblock {\em International Conference on Learning Representations}.

\bibitem[\protect\citeauthoryear{LeCun \bgroup et al.\egroup
  }{1998}]{lecun1998gradient}
LeCun, Y.; Bottou, L.; Bengio, Y.; and Haffner, P.
\newblock 1998.
\newblock Gradient-based learning applied to document recognition.
\newblock {\em Proc. of the IEEE} 86(11).

\bibitem[\protect\citeauthoryear{Luo}{2020}]{luo2020reinforcement}
Luo, J.
\newblock 2020.
\newblock {\em Reinforcement Learning for Generative Art}.
\newblock University of California, Santa Barbara.

\bibitem[\protect\citeauthoryear{Mariansky}{2020}]{mariansky2020transfer}
Mariansky, M.
\newblock 2020.
\newblock Transfer learning {StyleGAN} from ffhq faces to beetles is super
  weird.
\newblock \url{https://twitter.com/mmariansky/status/1226756838613491713}.
\newblock Accessed: 2021-02-04.

\bibitem[\protect\citeauthoryear{McCallum and
  Yee-King}{2020}]{mccallum2020network}
McCallum, L., and Yee-King, M.
\newblock 2020.
\newblock Network bending neural vocoders.
\newblock {\em NeurIPS 2020 Workshop on Machine Learning for Creativity and
  Design}.

\bibitem[\protect\citeauthoryear{Mordvintsev \bgroup et al.\egroup
  }{2020}]{mordvintsev2020growing}
Mordvintsev, A.; Randazzo, E.; Niklasson, E.; and Levin, M.
\newblock 2020.
\newblock Growing neural cellular automata.
\newblock {\em Distill} 5(2):e23.

\bibitem[\protect\citeauthoryear{Mouret and
  Clune}{2015}]{mouret2015illuminating}
Mouret, J.-B., and Clune, J.
\newblock 2015.
\newblock Illuminating search spaces by mapping elites.
\newblock {\em arXiv preprint arXiv:1504.04909}.

\bibitem[\protect\citeauthoryear{Park}{2020}]{park2020generating}
Park, S.-w.
\newblock 2020.
\newblock Generating novel glyph without human data by learning to communicate.
\newblock {\em NeurIPS 2020 Workshop on Machine Learning For Creativity and
  Design}.

\bibitem[\protect\citeauthoryear{Pinkney and
  Adler}{2020}]{pinkney2020interpolation}
Pinkney, J. N.~M., and Adler, D.
\newblock 2020.
\newblock Resolution dependent {GAN} interpolation for controllable image
  synthesis between domains.
\newblock {\em NeurIPS 2020 Workshop on Machine Learning for Creativity and
  Design}.

\bibitem[\protect\citeauthoryear{Pinkney}{2020a}]{pinkney2020aligned}
Pinkney, J. N.~M.
\newblock 2020a.
\newblock Aligned ukiyo-e faces dataset.
\newblock \url{https://www.justinpinkney.com/ukiyoe-dataset/}.
\newblock Accessed: 2021-06-28.

\bibitem[\protect\citeauthoryear{Pinkney}{2020b}]{pinkney2020awesome}
Pinkney, J. N.~M.
\newblock 2020b.
\newblock Awesome pretrained {StyleGAN2}.
\newblock \url{https://github.com/justinpinkney/awesome-pretrained-stylegan2}.
\newblock Accessed: 2020-02-05.

\bibitem[\protect\citeauthoryear{Pinkney}{2020c}]{pinkney2020matlab}
Pinkney, J. N.~M.
\newblock 2020c.
\newblock {MATLAB StyleGAN} playground.
\newblock \url{https://www.justinpinkney.com/matlab-stylegan/}.
\newblock Accessed: 2021-02-05.

\bibitem[\protect\citeauthoryear{Pouliot}{2020}]{pouliot2020gan}
Pouliot, A.
\newblock 2020.
\newblock {GAN} bending.
\newblock \url{https://darknoon.com/2020/03/03/gan-hacking/}.
\newblock Accessed: 2021-03-27.

\bibitem[\protect\citeauthoryear{Radford \bgroup et al.\egroup
  }{2021}]{radford2021learning}
Radford, A.; Kim, J.~W.; Hallacy, C.; Ramesh, A.; Goh, G.; Agarwal, S.; Sastry,
  G.; Askell, A.; Mishkin, P.; Clark, J.; et~al.
\newblock 2021.
\newblock Learning transferable visual models from natural language
  supervision.
\newblock {\em arXiv preprint arXiv:2103.00020}.

\bibitem[\protect\citeauthoryear{Ramesh \bgroup et al.\egroup
  }{2021}]{ramesh2021zero}
Ramesh, A.; Pavlov, M.; Goh, G.; Gray, S.; Voss, C.; Radford, A.; Chen, M.; and
  Sutskever, I.
\newblock 2021.
\newblock Zero-shot text-to-image generation.
\newblock {\em arXiv preprint arXiv:2102.12092}.

\bibitem[\protect\citeauthoryear{Rezende, Mohamed, and
  Wierstra}{2014}]{rezende2014stochastic}
Rezende, D.~J.; Mohamed, S.; and Wierstra, D.
\newblock 2014.
\newblock Stochastic backpropagation and approximate inference in deep
  generative models.
\newblock In {\em Proc. 31st International Conference on Machine Learning}.

\bibitem[\protect\citeauthoryear{Ritchie}{2007}]{ritchie2007some}
Ritchie, G.
\newblock 2007.
\newblock Some empirical criteria for attributing creativity to a computer
  program.
\newblock {\em Minds and Machines} 17(1):67--99.

\bibitem[\protect\citeauthoryear{Rumelhart, Hinton, and
  Williams}{1985}]{rumelhart1985learning}
Rumelhart, D.~E.; Hinton, G.~E.; and Williams, R.~J.
\newblock 1985.
\newblock Learning internal representations by error propagation.
\newblock Technical report, California Univ San Diego La Jolla Inst for
  Cognitive Science.

\bibitem[\protect\citeauthoryear{Růžička}{2020}]{ruzika2020gan}
Růžička, V.
\newblock 2020.
\newblock {GAN} explorer.
\newblock \url{https://github.com/previtus/GAN_explorer}.
\newblock Accessed: 2020-12-17.

\bibitem[\protect\citeauthoryear{Saleh and Elgammal}{2016}]{saleh2016large}
Saleh, B., and Elgammal, A.
\newblock 2016.
\newblock Large-scale classification of fine-art paintings: Learning the right
  metric on the right feature.
\newblock {\em International Journal for Digital Art History}.

\bibitem[\protect\citeauthoryear{Sarin}{2018}]{sarin2018playing}
Sarin, H.
\newblock 2018.
\newblock Playing a game of {GAN}struction.
\newblock \url{https://thegradient.pub/playing-a-game-of-ganstruction/}.
\newblock Accessed: 2020-12-15.

\bibitem[\protect\citeauthoryear{Saunders}{2019}]{saunders2019multi}
Saunders, R.
\newblock 2019.
\newblock Multi-agent-based models of social creativity.
\newblock In {\em Computational Creativity}. Springer.
\newblock  305--326.

\bibitem[\protect\citeauthoryear{Schultz}{2020a}]{schultz2020mixed}
Schultz, D.
\newblock 2020a.
\newblock Demo: How to mix models in {StyleGAN2}.
\newblock \url{https://www.youtube.com/watch?v=kbRkznsv9dk}.
\newblock Accessed: 2020-02-07.

\bibitem[\protect\citeauthoryear{Schultz}{2020b}]{schutlz2020you}
Schultz, D.
\newblock 2020b.
\newblock You {A}re {H}ere.
\newblock
  \url{https://artificial-images.com/project/you-are-here-machine-learning-film/}.
\newblock Accessed: 2021-06-28.

\bibitem[\protect\citeauthoryear{Schultz}{2021}]{schultz2021personal}
Schultz, D.
\newblock 2021.
\newblock Personal communication.

\bibitem[\protect\citeauthoryear{Shaker}{2016}]{shaker2016intrinsically}
Shaker, N.
\newblock 2016.
\newblock Intrinsically motivated reinforcement learning: A promising framework
  for procedural content generation.
\newblock In {\em 2016 IEEE Conference on Computational Intelligence and Games
  (CIG)},  1--8.
\newblock IEEE.

\bibitem[\protect\citeauthoryear{Shane}{2020}]{shane2020cat}
Shane, J.
\newblock 2020.
\newblock Trained a neural net on my cat and regret everything.
\newblock
  \url{https://aiweirdness.com/post/615654447163621376/trained-a-neural-net-on-my-cat-and-regret}.
\newblock Accessed: 2020-02-05.

\bibitem[\protect\citeauthoryear{Simon}{2019}]{simon2019dimensions}
Simon, J.
\newblock 2019.
\newblock Dimensions of dialogue.
\newblock \url{https://www.joelsimon.net/dimensions-of-dialogue.html}.
\newblock Accessed: 2020-12-15.

\bibitem[\protect\citeauthoryear{Sloan}{2012}]{sloan2012flipflop}
Sloan, R.
\newblock 2012.
\newblock Dancing the flip flop.
\newblock \url{https://www.robinsloan.com/notes/flip-flop/}.
\newblock Accessed: 2021-03-27.

\bibitem[\protect\citeauthoryear{Som}{2020}]{som2020strange}
Som, M.
\newblock 2020.
\newblock {Strange Fruit}.
\newblock
  \url{http://www.aiartonline.com/highlights-2020/mal-som-errthangisalive/}.
\newblock Accessed: 2021-02-05.

\bibitem[\protect\citeauthoryear{Som}{2021}]{som2021personal}
Som, M.
\newblock 2021.
\newblock Personal communication.

\bibitem[\protect\citeauthoryear{Thanh-Tung and
  Tran}{2020}]{thanh2020catastrophic}
Thanh-Tung, H., and Tran, T.
\newblock 2020.
\newblock Catastrophic forgetting and mode collapse in {GANs}.
\newblock In {\em Proc. International Joint Conference on Neural Networks
  (IJCNN)}.

\bibitem[\protect\citeauthoryear{Wang \bgroup et al.\egroup
  }{2020a}]{wang2020enhanced}
Wang, R.; Lehman, J.; Rawal, A.; Zhi, J.; Li, Y.; Clune, J.; and Stanley, K.
\newblock 2020a.
\newblock Enhanced poet: Open-ended reinforcement learning through unbounded
  invention of learning challenges and their solutions.
\newblock In {\em Proc. International Conference on Machine Learning}.

\bibitem[\protect\citeauthoryear{Wang \bgroup et al.\egroup
  }{2020b}]{wang2020towards}
Wang, X.; Bylinskii, Z.; Hertzmann, A.; and Pepperell, R.
\newblock 2020b.
\newblock Towards quantifying ambiguities in artistic images.
\newblock {\em ACM Trans. Appl. Percept.}

\bibitem[\protect\citeauthoryear{White}{2016}]{white2016sampling}
White, T.
\newblock 2016.
\newblock Sampling generative networks.
\newblock {\em arXiv preprint arXiv:1609.04468}.

\bibitem[\protect\citeauthoryear{Zhu \bgroup et al.\egroup
  }{2017}]{zhu2017unpaired}
Zhu, J.-Y.; Park, T.; Isola, P.; and Efros, A.~A.
\newblock 2017.
\newblock Unpaired image-to-image translation using cycle-consistent
  adversarial networks.
\newblock In {\em Proc. IEEE international conference on computer vision}.

\end{thebibliography}

\end{document}